\documentclass{article}
\usepackage{spconf,amsmath,graphicx}
\usepackage{amsfonts} 
\usepackage{etoolbox} 
\usepackage{booktabs} 
\usepackage[noend]{algpseudocode} 
\usepackage{algorithmicx,algorithm} 
\usepackage{multirow} 
\usepackage{amsmath} 
\usepackage{bbding} 
\usepackage{caption} 
\usepackage{subfigure} 
\usepackage{color} 
\usepackage{url} 
\usepackage{hyperref} 
\usepackage{graphicx} 
\usepackage[bottom]{footmisc} 

\usepackage{xspace}
\newcommand{\method}{\textbf{\textsc{CLseg}}\xspace}

\title{\method: Contrastive Learning of Story Ending Generation}
\name{Yuqiang Xie \quad Yue Hu\sthanks{Corresponding author. E-mail: huyue@iie.ac.cn}	\quad Luxi Xing \quad Yunpeng Li \quad Wei Peng  \quad Ping Guo}
\address{Institute of Information Engineering, Chinese Academy of Sciences, Beijing, China \\
	School of Cyber Security, University of Chinese Academy of Sciences, Beijing, China}
\begin{document}
	%
	\maketitle
	\renewcommand{\thefootnote}{\fnsymbol{footnote}}
	\begin{abstract}
		Story Ending Generation (SEG) is a challenging task in natural language generation. 
		Recently, methods based on Pre-trained Language Models (PLM) have achieved great prosperity, which can produce fluent and coherent story endings. 
		However, the pre-training objective of PLM-based methods is unable to model the consistency between story context and ending. 
		The goal of this paper is to adopt contrastive learning to generate endings more consistent with story context, while there are two main challenges in contrastive learning of SEG. 
		First is the negative sampling of wrong endings inconsistent with story contexts. 
		The second challenge is the adaptation of contrastive learning for SEG. 
		To address these two issues, we propose a novel \textbf{C}ontrastive \textbf{L}earning framework for \textbf{S}tory \textbf{E}nding \textbf{G}eneration (\method)\footnotemark[2]\footnotetext[2]{Code and Data: \url{https://github.com/IndexFziQ/CLSEG}}, which has two steps: multi-aspect sampling and story-specific contrastive learning. 
		Particularly, for the first issue, we utilize novel multi-aspect sampling mechanisms to obtain wrong endings considering the consistency of order, causality, and sentiment. 
		To solve the second issue, we well-design a story-specific contrastive training strategy that is adapted for SEG.
		Experiments show that \textbf{\textsc{Clseg}} outperforms baselines and can produce story endings with stronger consistency and rationality.
	\end{abstract}
	\section{Introduction}
	\label{sec:intro}
	
	Story Ending Generation (SEG) is a challenging task in natural language generation and artificial intelligence (AI), which aims to complete the plot and conclude a story given a story context \cite{Mostafazadeh2016ACA,Sharma2018TacklingTS,Zhao2018FromPT}. 
	SEG systems require to understand the story context, and then generate coherent, reasonable, and diversified endings according to the temporal and causal relationships. 
	Previous state-of-the-art Seq2Seq based methods, like GPT-2 \cite{Radford2019LanguageMA} and BART \cite{DBLP:conf/acl/LewisLGGMLSZ20}, mainly generate in a left-to-right manner and train with language model objective, which undergo an issue of generating inconsistent and safe endings. 
	
	To solve this issue, researches consider integrating semantic \cite{Huang2021StoryEG,Xu2020ControllableSG}, commonsense \cite{Li2018GeneratingRA,Guan2019StoryEG,Paul2021COINSDG}, sentiment/emotion \cite{Luo2019LearningTC,Brahman2020ModelingPE} or even multi-modal \cite{Huang2021IgSEGIS} knowledge into backbone models. 
	Particularly, \cite{Guan2020AKP} introduces a self-supervised task to distinguish true stories from auto-constructed fake stories to incorporate commonsense knowledge into GPT-2, so that the content of generated endings appears more coherent. 
	In terms of consistency and rationality, however, there is still a big gap between machines and humans. 
	One reason is that the proposed self-supervised classification task can learn the  difference between true story and fake story, but it can not directly model the consistency between story context and endings.
	
	We argue that the comparison of story context and various endings is vital to generating more consistent story endings, therefore it is necessary to seek an approach to better grasp the consistency. 
	Based on the above analysis, we adopt the paradigm of contrastive learning \cite{Arora2019ATA} to introduce an approach for story ending generation which achieves the goal of modeling the consistency between story context and candidate endings.
	\textbf{There are two crucial problems in contrastive learning of story ending generation:}
	
	\textbf{Q1:} \textit{How to generate high-quality wrong endings inconsistent with story contexts?} 
	
	\textbf{Q2:} \textit{How to design a contrastive training strategy adapted for story ending generation?}
	
	To address the aforementioned problems, we propose a novel \textbf{C}ontrastive \textbf{L}earning framework for \textbf{S}tory \textbf{E}nding \textbf{G}eneration (\method), which has two steps: multi-aspect sampling and story-specific contrastive learning. 
	Targeted at \textbf{Problem Q1}, multi-aspect sampling mechanisms are utilized to sample high-quality wrong endings by considering the consistency of order, causality, and sentiment. 
	To solve \textbf{Problem Q2}, a story-specific contrastive training strategy is designed, which is adapted for SEG.  
	Experimental results demonstrate that \method outperforms baselines and can produce more consistent and reasonable story endings.
	
	Our contributions are summarized as below: 
	\textbf{(1)} We propose a novel \method framework for story ending generation that can generate endings more consistent to the story context. 
	\textbf{(2)} To achieve contrastive learning of SEG, we introduce novel multi-aspect sampling mechanisms and a story-specific contrastive training strategy. 
	\textbf{(3)} Automatic and manual evaluations demonstrate that \method outperforms baselines on recall-oriented metrics and can generate story endings with stronger consistency and rationality.

	\section{\method}
	\label{sec:method}
	
	\subsection{Overview}
	SEG task can be formulated as follows: given a story context consisting of a sequence of sentences $ \boldsymbol{\mathcal{X}} = \{\mathcal{X}_1, \mathcal{X}_2, \dots , \mathcal{X}_m\}$ where $m=4$ in this paper. 
	$\mathcal{X}_i = \{x_i^1, x_i^2, ..., x_i^n\}$ represents the $i$-th sentence that consists of $n$ words, and $1\le i \le m$.
	The goal of SEG is to generate a story ending $\boldsymbol{\mathcal{Y}}$ according to the given story context $\boldsymbol{\mathcal{X}}$.
	
	\method is based on a LM architecture GPT-2 \cite{Radford2019LanguageMA}. 
	Firstly, aiming at generating candidate wrong endings, we design a series of methods considering the consistency of story order, causality, and sentiment in the multi-aspect sampling.
	Specifically, we introduce three novel negative sampling mechanisms: shuffled ordering, counterfactual rewriting and reverse sentiment. 
	Then, we design a story-specific contrastive training strategy to further pre-train GPT-2 to learn the consistency between story context and endings, which compares right ending and wrong endings, thus helping GPT-2 to generate a more consistent ending.
	
	\begin{figure}[t]
		\centering
		\includegraphics[width=8.2cm]{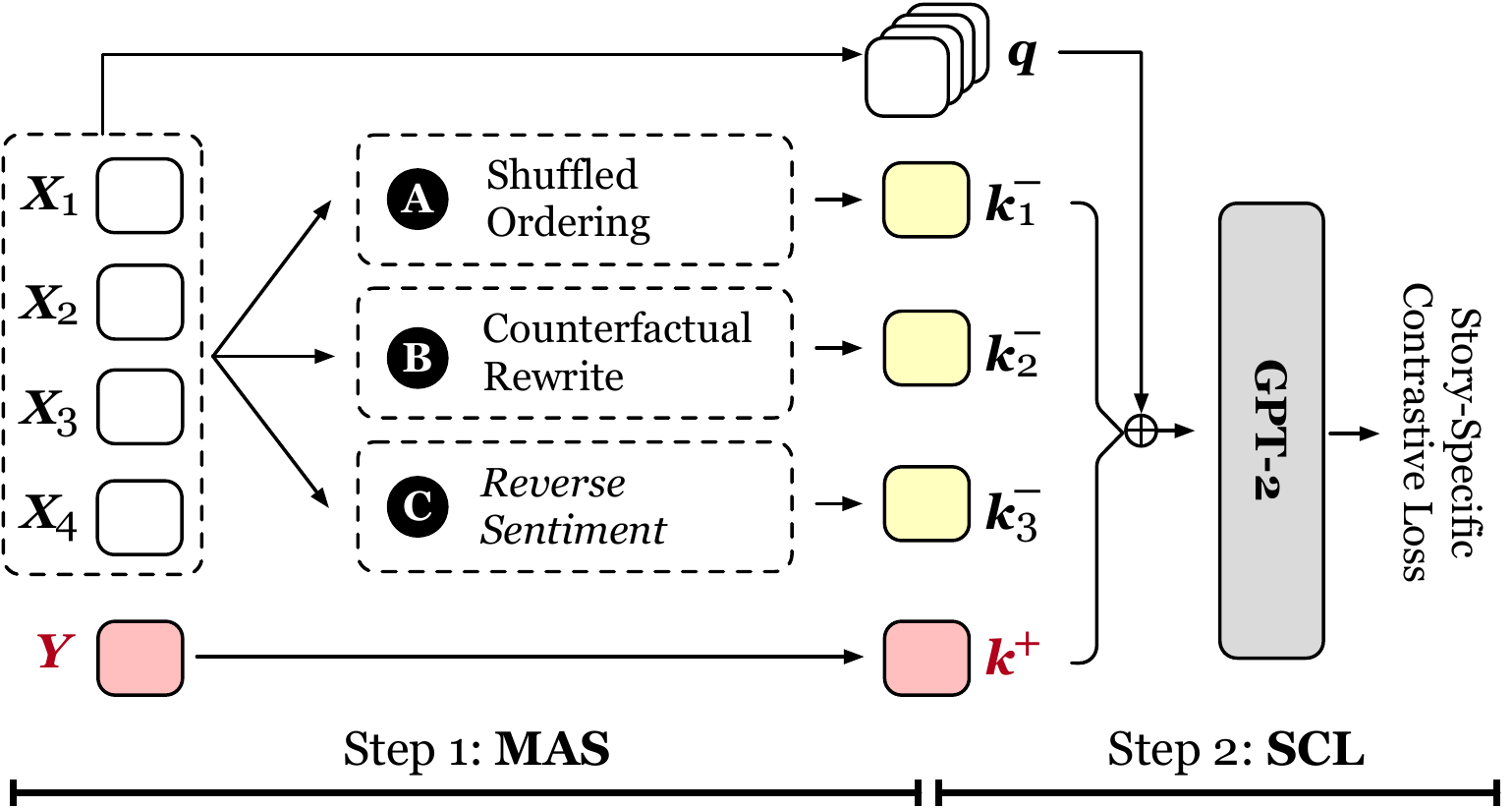}
		\caption{Overview of \method.}
		\label{fig:overview}
	\end{figure}

	\subsection{Multi-Aspect Sampling (MAS)}
	\label{sec:MAS}
	
	In order to address \textbf{Problem Q1}, we generate high-quality wrong endings considering three aspects: order, causality, and sentiment. 
	As exhibited in Fig. \ref{fig:overview}, Multi-Aspect Sampling (\textbf{MAS}) consists of three sampling mechanisms to generate wrong endings $\{\boldsymbol{\mathcal{Y}^-_{\text{SO}}}, \boldsymbol{\mathcal{Y}^-_{\text{CR}}}, \boldsymbol{\mathcal{Y}^-_{\text{RS}}}\}$ based on the original story context $\boldsymbol{\mathcal{X}}$. 
	Specifically, Shuffled Ordering (SO) is used for generating temporal inconsistent endings. 
	Counterfactual Rewriting (CR) is designed for creating causal inconsistent endings.
	Reverse Sentiment (RS) is utilized for writing sentiment inconsistent endings.
	
\begin{figure}[t]
		\centering
		\includegraphics[width=7cm]{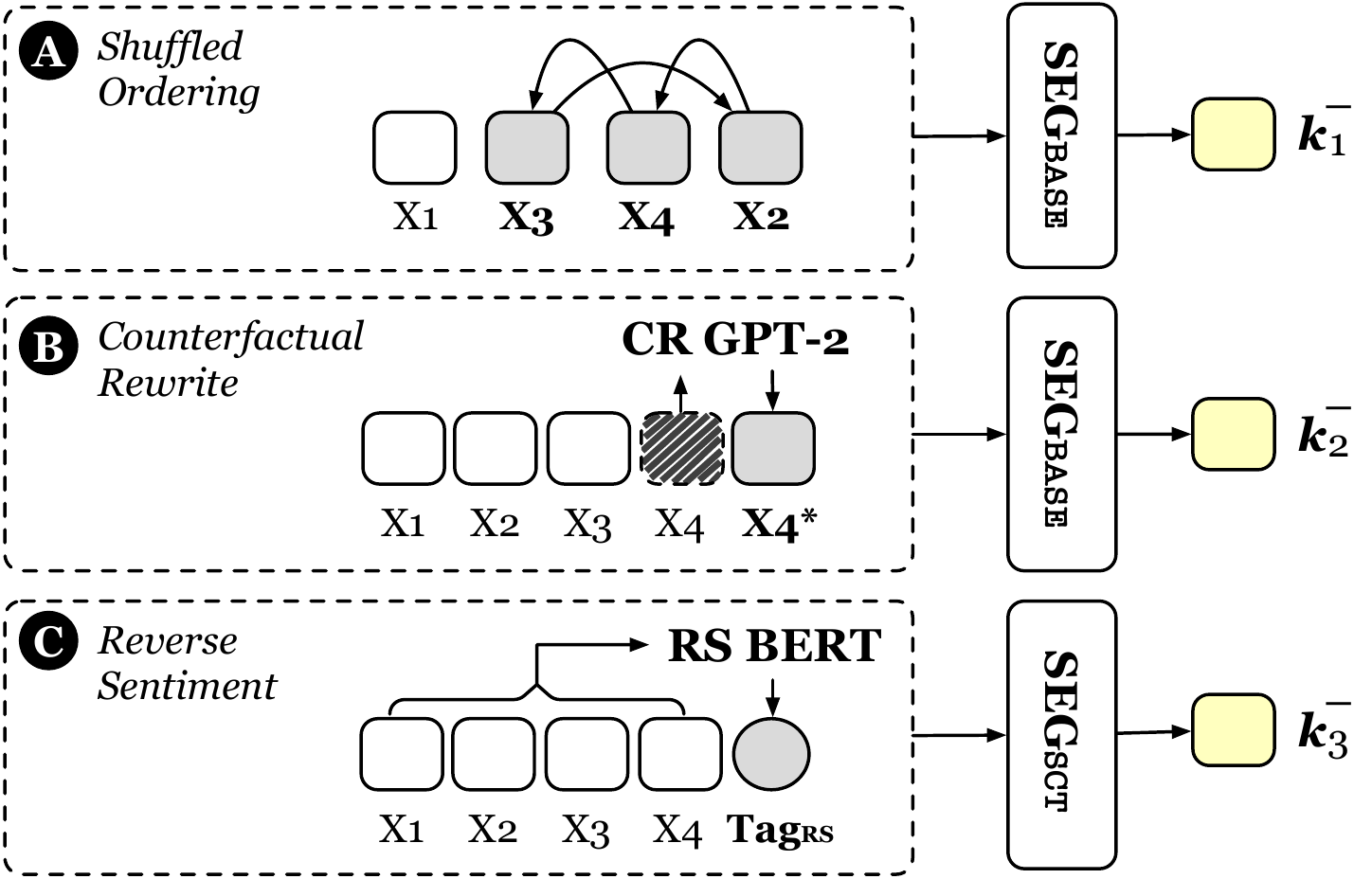}
		\caption{Details of Multi-Aspect Sampling.}
		\label{fig:mas}
	\end{figure}

	\noindent\textbf{Shuffled Ordering (SO)}
	For the purpose of modeling the temporal consistency between story context and ending, we design a SO mechanism to generate wrong endings, which is similar to \cite{Guan2020AKP,ksem2020xie}. 
	In the beginning, we fine-tune a story ending generation model SEG{\small{\texttt{BASE}}} based on GPT-2 \cite{Radford2019LanguageMA} on ROCStories \cite{Mostafazadeh2016ACA}.
	As shown in Fig. \ref{fig:mas} (A), we randomly shuffle the order of story context. 
	The wrong ending $\boldsymbol{\mathcal{Y}^-_{\text{SO}}}$ generated by SO mechanism is defined as:
	\begin{equation}
		\boldsymbol{\mathcal{Y}^-_{\text{SO}}}= \text{SEG{\small{\texttt{BASE}}}}(\textit{shuffled}(\{\mathcal{X}_1, \mathcal{X}_2, \dots , \mathcal{X}_m\}))
	\end{equation}
	The generated wrong ending will be considered as the wrong ending for the next contrastive learning step.
	
	\noindent\textbf{Counterfactual Rewriting (CR)}
	Aiming at taking the causal consistency between story context and ending into account, we introduce a Counterfactual Rewriting (CR) mechanism to create wrong endings.
	As demonstrated in Fig. \ref{fig:mas} (B), we replace the last event $\mathcal{X}_m$ of story context with a counterfactual rewriting event $\mathcal{X}^*_m$. 
	Thus, the wrong ending $\boldsymbol{\mathcal{Y}^-_{\text{CR}}}$ generated by CR mechanism as below:
	\begin{equation}
		\boldsymbol{\mathcal{Y}^-_{\text{CR}}}= \text{SEG{\small{\texttt{BASE}}}}(\{\mathcal{X}_1, \mathcal{X}_2, \dots , \mathcal{X}^*_m\})
	\end{equation}
	where $\mathcal{X}^*_m$ is generated by \textbf{CR GPT-2} that trained on \textsc{TimeTravel} \cite{Qin2019CounterfactualSR}. \textsc{TimeTravel} is a large-scale human-annotated counterfactual story generation data set which consists of almost 100K Counterfactual Rewriting sentence pairs (original event, counterfactual event). 
	$\boldsymbol{\mathcal{Y}^-_{\text{CR}}}$ will be given to the next contrastive learning step.

	\begin{table*}[t]
		\centering
		\scalebox{0.8}{
			\begin{tabular}{lccccccccccc}
				\toprule[1pt]
				\textbf{Models}  & \textbf{BLEU} $\uparrow$   & \textbf{R-1-P} $\uparrow$  & \textbf{R-1-R} $\uparrow$   & \textbf{R-1-F1} $\uparrow$ & \textbf{R-2-P} $\uparrow$  & \textbf{R-2-R} $\uparrow$   & \textbf{R-2-F1} $\uparrow$ & \textbf{R-L-P} $\uparrow$  & \textbf{R-L-R} $\uparrow$   & \textbf{R-L-F1} $\uparrow$& \textbf{Meteor} $\uparrow$   \\
				\toprule[1pt]
				\textbf{GPT-2(PT)} &   1.14         &   13.99         &    \textbf{15.99}        &   13.56       &   1.34    &  \textbf{2.51}  &  1.68          &  13.17   & \textbf{15.89}   &   12.65         &    10.36      \\
				\textbf{GPT-2(FT)} &  \textbf{2.57}         &     15.15       &      12.55      &    13.30        &  \textbf{2.12}         &  1.82  &      1.87      & 13.87     &  11.45  &      12.14      & 10.48       \\
				\textbf{GPT-2(GCL)} &   1.71         &   14.81        & 14.66        &  14.21        & 1.82   &   2.13           &  1.89          & 13.67   &  12.56  &  13.12         &   10.54       \\
				\toprule[0.5pt]
				\method &   1.97           &      \textbf{15.68}    &   14.73       &     \textbf{14.63}      & 1.87   &      2.15         &  \textbf{1.91}        & \textbf{14.44}     &  13.61  &      \textbf{13.48}      & \textbf{10.73}        \\
				\toprule[1pt]
		\end{tabular}}
		\caption{The results of automatic evaluation considering common-used metrics. $\uparrow$ indicates the higher, the better.}
		\label{table-gen}
	\end{table*}

	\noindent\textbf{Reverse Sentiment (RS)}
	In order to consider the sentiment consistency between story context and ending, we propose a Reverse Sentiment (RS) mechanism to writing wrong endings. As displayed in Fig. \ref{fig:mas} (C), the wrong ending $\boldsymbol{\mathcal{Y}^-_{\text{RS}}}$ generated by RS mechanism is defined as:
	\begin{equation}
		\boldsymbol{\mathcal{Y}^-_{\text{RS}}}= \text{SEG{\small{\texttt{SCT}}}}(\{\mathcal{X}_1, \mathcal{X}_2, ..., \mathcal{X}_m, \text{Tag}_\text{RS}\})
	\end{equation}
	where a reverse sentiment tag $\text{Tag}_\text{RS} \in \{0,1\}$ is predicted by an \textbf{RS BERT} which is a BERT{\small{\texttt{LARGE}}} \cite{DBLP:conf/naacl/DevlinCLT19} fine-tuned on the SST-2 \cite{Socher2013RecursiveDM} dataset.
	We utilize the \textbf{RS BERT} to label Story Cloze Test (SCT) \cite{Mostafazadeh2016ACA} and extract the examples that the sentiment orientation of wrong ending is opposite to the sentiment orientation of story context. 
	In statistics, 86.5\% of examples fit our setting. 
	A story ending generation model SEG{\small{\texttt{SCT}}} is a GPT-2 \cite{Radford2019LanguageMA} fine-tuned on labeled SCT \cite{Mostafazadeh2016ACA}.
	$\boldsymbol{\mathcal{Y}^-_{\text{RS}}}$ will be fed to the next contrastive learning step.

	\subsection{Story-Specific Contrastive Learning (SCL)}
	\label{sec:MSCL}
	
	To solve \textbf{Problem Q2}, Story-Specific Contrastive Learning (\textbf{SCL}) is designed, which is adapted for SEG.
	As demonstrated in Fig. \ref{fig:overview}, we design a story-specific contrastive training strategy for \method.  
	Different from contrastive learning for natural language understanding tasks (pre-training encoder), we focus on pre-training decoder for SEG.
	
	In Section \ref{sec:MAS}, we explicitly construct negative examples for each right endings in ROCStories \cite{Mostafazadeh2016ACA} for \textbf{SCL}. 
	We encourage story context $\boldsymbol{\mathcal{X}}$ (the query, denoted as $\boldsymbol{q}$) to be as consistent as possible to the right ending $\boldsymbol{\mathcal{Y}}$ (the positive sample, denoted as $\boldsymbol{k}^+$) but inconsistent to all wrong endings $\{\boldsymbol{\mathcal{Y}^-_{\text{SO}}}, \boldsymbol{\mathcal{Y}^-_{\text{CR}}}, \boldsymbol{\mathcal{Y}^-_{\text{RS}}}\}$ based on the original story context $\boldsymbol{\mathcal{X}}$
	(i.e., the negative sample with inconsistent order, causal, and sentiment denoted as $\{\boldsymbol{k}^-_i\}^3_{i=1}=\{\boldsymbol{k}^-_1, \boldsymbol{k}^-_2, \boldsymbol{k}^-_3\}$ individually) in a training batch. 
	Inspired by \cite{Welleck2020NeuralTG}, the story-specific contrastive loss for $\boldsymbol{\mathcal{X}}$ is defined as:
	\begin{equation}
		\begin{aligned}
			\mathcal{L}_{\text {SCL }}^{t} \left(p_{\theta}\left(\boldsymbol{\mathcal{Y}}|\boldsymbol{\mathcal{X}}\right), \{\boldsymbol{\mathcal{Y}}^{-}_i\}^3_{i=1}\right) =-\log p_{\theta}\left(\boldsymbol{k}^+_{t} \mid \boldsymbol{k}^+_{<t},\boldsymbol{q} \right) \\ -\alpha \cdot \frac{1}{N} \sum^N_{i} \log \left(1-p_{\theta}\left(\boldsymbol{k}^-_{i,t} \mid \boldsymbol{k}^-_{i,<t},\boldsymbol{q} \right)\right) \\
		\end{aligned}
	\end{equation}
	where $p_{\theta}$ represents backbone LM GPT-2, $t$ is the time step, and $\alpha$ is a hyper-parameter. In this paper, we set $N=3$.

	\section{Experiments}
	\label{sec:Experiments}
	
	\subsection{Experimental Setup}
	\noindent\textbf{Dataset}
	We evaluate our model on the ROCStories \cite{Mostafazadeh2016ACA} corpus. The corpus contains 98,162 five-sentence stories for evaluating story understanding and generation. 
	Following the 8:1:1 splitting ratio, we obtain 78,530/9,816/9,816 five-sentence stories as train/dev/test sets for the SEG task.
	
	\noindent\textbf{Implement Details}
	The stories are tokenized using byte pair encoding (BPE). 
	We set the parameters following the medium version of \cite{Radford2019LanguageMA}’s design. 
	The batch size is 32 during fine-tuning on the ROCStories corpus using Adam optimizer with an initial learning rate of 5e-5. We set $\alpha=1$.
	We generate stories using a greedy sampling strategy. 
	We use the HuggingFace\footnotemark[3]
	\footnotetext[3]{\url{https://github.com/huggingface/transformers}} \cite{DBLP:conf/emnlp/WolfDSCDMCRLFDS20}
	PyTorch \cite{DBLP:conf/nips/PaszkeGMLBCKLGA19} implementation on Tesla V100 GPU.
	
	\subsection{Evaluation Metrics}
	\label{ssec:Metrics}
	\noindent\textbf{Automatic}
	We use the following metrics for automatic evaluation\footnotemark[5]\footnotetext[5]{Evaluate with \url{https://github.com/thu-coai/OpenMEVA} } \cite{Guan2021OpenMEVAAB}:
	\textbf{(1) BLEU} \cite{papineni2002bleu} is used for evaluating the overall quality of the generated story. We use geometric mean of 1-gram to 4-gram.
	\textbf{(2) ROUGE (R-n-P/R/F1)} \cite{Lin2004ROUGEAP} (with n=1, 2, L) is used to measure the recall-oriented similarity between automatically generated and reference results, including Precision, Recall and F1.
	\textbf{(3) Meteor} \cite{banerjee2005meteor} is based on the harmonic mean of unigram precision and recall, with recall weighted higher than precision. 
	This metric can produce good correlation with human judgement.

	 \begin{table}[t]
	 	\centering
	 	\scalebox{0.78}{
	 		\begin{tabular}{lccccc}
	 			\toprule[1pt]
	 			\multirow{2}{*}{\textbf{Model}} & \multicolumn{2}{c}{\textbf{Quality} $\uparrow$} & \multicolumn{3}{c}{\textbf{Consistent} $\uparrow$} \\
	 			& \textbf{Fluency}      & \textbf{Coherence}    
	 			& \textbf{Order}          & \textbf{Causal}   & \textbf{Sentiment}   \\
	 			\toprule[1pt]
	 			\textbf{GPT-2(PT)}     &     1.87      &   1.58          &    1.22          & 1.07 &  1.19   \\
	 			\textbf{GPT-2(FT)}    &      \textbf{2.24}      &  2.13             &  1.47             & 1.67 & 1.68 \\
	 			\textbf{GPT-2(GCL)}    &     2.08       &   2.07            &   1.62          &   1.73 &  1.63   \\
	 			\toprule[0.5pt]
	 			\method   &      2.20       &   \textbf{2.43}            &  \textbf{2.08}            &  \textbf{2.19}    &  \textbf{2.02}   \\
	 			\toprule[1pt]
	 	\end{tabular}}
	 	\caption{Manual Evaluation in terms of quality and rationality about the generated story endings.}
	 	\label{table-humaneval}
	 \end{table}
	
	\noindent\textbf{Manual}
	We also conduct a manual evaluation of generated story endings. Crowd-workers are required to evaluate actions on a 0-3 scale (3 being very good) from two different perspectives: 
	\textbf{(1) Content Quality} to indicate whether the generated story ending is fluent and coherent. 
	\textbf{(2) Content Rationality} to assess whether the story endings are reasonable and consistent to the story context.

	\subsection{Results}
	\noindent\textbf{Baselines}
	We use the following baselines: 
	\textbf{GPT-2(PT):} This model directly use the public checkpoint of pre-trained parameters for story ending generation which follows the setting of paper \cite{Radford2019LanguageMA}.
	\textbf{GPT-2(FT):} This model is fine-tuned on the ROCStories corpus from the public checkpoint of pre-trained parameters.
	\textbf{GPT-2(GCL):} This model is pre-trained similar to our model, but the negative sampling is in general noisy ways: \textit{random shuffle/drop/replace tokens}.
	
	\noindent\textbf{Automatic Evaluation}
	The results of automatic evaluation are shown in Table \ref{table-gen}. 
	GPT-2(PT) shows highest recall but the lowest precision and BLEU scores. 
	One reason is that GPT-2(PT) tends to generate as many high-frequency co-occurring tokens as possible. 
	However, GPT-2(PT)'s generated endings is of low quality. \method outperforms GPT-2(FT) and GPT-2(GCL) in terms of recall-oriented Meteor, and the recall of Rouge scores than all the baselines, indicating better consistency with the reference golden endings. 
	Meanwhile, compared with GPT-2(PT), the content quality (BLEU and Precision of Rouge) of \method's generated endings is guaranteed.
	Besides, we have conducted p-value \cite{sogaard-etal-2014-whats} evaluation for significance of the differences. 
	The results show that our method outperforms all baseline models significantly with p-value $< 0.012$ (Wilcoxon signed-rank test).
	
	\noindent\textbf{Manual Evaluation}
	We perform a manual evaluation between our model and baselines. We randomly generate 200 stories from the test set. 
	For each story, we hire three annotators to give a score in terms of content quality (fluency, coherence) and content rationality (order, causal, sentiment). 
	For each aspect, we use average of the three annotations. We adopt majority voting to make the final decisions among the annotators. 
	As exhibited in Table \ref{table-humaneval}, all the results show that our model outperforms baselines significantly in content coherence, and all aspects of consistency.

	\section{Discussion and Analysis}
	\label{sec:DA}

	\noindent\textbf{Ablation Study}
	An ablation study is conducted on the ROCStories dataset to examine the impact of each negative sampling mechanism separately. 
	We train the model each time by using one of our negative sampling mechanisms. 
	As shown in Table \ref{table-ablation}, the results (ROUGE recall accuracy and Meteor) illustrate the harms that the elimination of each of the proposed negative sampling mechanisms (SO, CR or RS) from \method could cause.
	In summary, all negative sampling mechanisms contribute for \method and RS shows best performance.

	\begin{table}[t]
		\centering
		\scalebox{0.8}{
			\begin{tabular}{lccccc}
				\toprule[1pt]
				\textbf{Model}  & \textbf{BLEU} $\uparrow$  & \textbf{R-1-R} $\uparrow$   & \textbf{R-2-R} $\uparrow$    &  \textbf{R-L-R} $\uparrow$  & \textbf{Meteor} $\uparrow$  \\
				\toprule[1pt]
				\method &    1.97        &     \textbf{14.73}       &      \textbf{2.15}     &    \textbf{13.61}       &  \textbf{10.73}     \\
				only SO &  1.94          &     13.77       &      1.96      &    12.78          &  10.16     \\
				only CR &  2.13          &    13.71       &      1.96      &    12.66           & 10.20      \\
				only RS &  \textbf{2.33}        &     14.24       &      2.03     &    13.02            & 10.64    \\
				\toprule[1pt]
		\end{tabular}}
		\caption{Ablation study of \method.}
		\label{table-ablation}
	\end{table}
	
	\begin{table}[t]
		\centering
		\scalebox{0.8}{
			\begin{tabular}{ll}
				\toprule[1pt]
				\toprule[1pt]
				\textbf{Context}     &  \multicolumn{1}{p{7.5cm}}{Yesterday, I needed to buy a belt so I went to the mall. After choosing which store to shop from, I quickly got out of my car. Upon going inside there was an associate to help me. We talked about what i wanted and she led me to the section.} \\
				\toprule[1pt]
				\textbf{Golden}   & \multicolumn{1}{p{7.5cm}}{I picked out a great belt and left feeling excellent!} \\
				\toprule[0.5pt]
				\textbf{GPT-2(PT)}   & \multicolumn{1}{p{7.5cm}}{I was so out of my car I was about to buy a belt.} \\
				\textbf{GPT-2(FT)}  & \multicolumn{1}{p{7.5cm}}{I was so quickly and quickly, i was so out of my car.}   \\
				\textbf{GPT-2(GCL)}  & \multicolumn{1}{p{7.5cm}}{I was so so so ...}   \\
				\method   & \multicolumn{1}{p{7.5cm}}{After I got my belt, I went to the store.}    \\
				\toprule[1pt]
				\toprule[1pt]
		\end{tabular}}
			\caption{Generated story endings by different models.}
			\label{table-case}
		\end{table}

	\noindent\textbf{Case Study}
	In this part, we present some generated examples in Table \ref{table-case}. 
	The first two lines are the story context and the golden right ending.
	From the next 4 lines, our model can generate more consistent and reasonable story ending to the story context than baselines.
	In specific, \textbf{GPT-2(PT)} generate a repetitive event to story context. 
	Besides, the generated ending of \textbf{GPT-2(FT)} are inconsistent and far from the golden endings.
	And, \textbf{GPT-2(GCL)} fails to generate a event. 
	The examples show that \method can generate more consistent endings by contrastive learning high-quality wrong endings.
	
	\noindent\textbf{Error Analysis}
	In order to explore the limitation of \method, we also perform error analysis by studying many inconsistent endings lost to baselines in manual evaluation.
    Hence, we manually examine all the bad endings in the pairwise comparison between our model and two strong baselines \textbf{GPT-2(FT/GCL)} to analyze the types of errors. Error types indicate the causes that influence the performance. 
	The numbers of endings lost to our model are 43/58 of 100/100 in total for \textbf{GPT-2(FT/GCL)}, respectively. 
	And there are 35 endings of 200 generated by \method losing to these two baselines.
	
	We conclude three main types of error from the lost endings: \textbf{Repetition} (repeating the same words or event to the story context), \textbf{Conflicting} (inconsistent to the story context), and \textbf{Ambiguous} (difficult to understand the ending). 
	As displayed in Table \ref{table-errorcase}, \method generates the repeated words corresponding to story context. 
	The last two examples show conflicting and ambiguous endings. 
	All in all, the analysis result illustrates that generating a high-quality consistent ending to the story context is still a challenging task.

	\begin{table}[t]
		\centering
		\scalebox{0.8}{
			\begin{tabular}{ll}
				\toprule[1pt]
				\toprule[1pt]
				\textbf{Error Type} & \textbf{Story Context + Ending} \\
				\toprule[1pt]
				\textbf{Repetition}     & \multicolumn{1}{p{8cm}}{Morgan enjoyed long \textit{walks} on \textit{the beach}. She and her boyfriend decided to go ... Morgan decided to \textit{propose} to her boyfriend. \textit{The walk to the beach and the propose.}}  \\
				\textbf{Conflicting}     &\multicolumn{1}{p{8cm}}{Frank had been drinking beer. He got a call from his girlfriend ...  \textit{Since Frank was already a bit drunk, he could not drive. Frank had to drive to his date.}} \\
				\textbf{Ambiguous}     &\multicolumn{1}{p{8cm}}{Sunny enjoyed going to the beach. As she stepped out of her car, ... Sunny got back into her car and heading towards the mall. \textit{She was going to the park.}} \\
				\toprule[1pt]
				\toprule[1pt]
		\end{tabular}}
		\caption{Typical errors generated by \method. \textit{Italic} words denote the error generated story endings.}
		\label{table-errorcase}
	\end{table}
	
	\section{Conclusion and Future Work}
	\label{sec:Conclusion}
	
	In this work, we propose a novel \method framework for story ending generation which has two steps: multi-aspect sampling (MAS) and story-specific contrastive learning (SCL) to generate endings more consistent to story context. 
	Experimental results demonstrate that \method outperforms baselines and creates more consistent and reasonable story endings.
	In the future, we will consider constructing harder yet high-quality wrong endings, designing more appropriate CL training strategies and applying \method to more NLP tasks.
	
	\section{ACKNOWLEDGMENTS}
    We thank all anonymous reviewers for their constructive comments. This work is supported by the National Natural Science Foundation of China (No.62006222 and No.U21B2009).



	\vfill\pagebreak

	\small
	\bibliographystyle{IEEEbib}
	\bibliography{refs}
	
\end{document}